\documentclass{article}
\usepackage{graphicx}
\usepackage{amsfonts}
\usepackage{amsmath}

\title{Graph Neural Network for Video Relocalization}
\author{Yuan~Zhou, Mingfei~Wang, Ruolin~Wang, Shuwei~Huo}
\begin{document}
	\maketitle

% \newpage
% \tableofcontents

% \newpage

\section{Abstract}

In this paper, we focus on video relocalization task, which uses a query video clip as input to retrieve a semantic relative video clip in another untrimmed long video. we find that in video relocalization datasets, there exists a phenomenon showing that there does not exist consistent relationship between feature similarity by frame and feature similarity by video, which affects the feature fusion among frames. However, existing video relocalization methods do not fully consider it. Taking this phenomenon into account, in this article, we treat video features as a graph by concatenating the query video feature and proposal video feature along time dimension, where each timestep is treated as a node, each row of the feature matrix is treated as feature of each node. Then, with the power of graph neural networks, we propose a Multi-Graph Feature Fusion Module to fuse the relation feature of this graph. After evaluating our method on ActivityNet v1.2 dataset and Thumos14 dataset, we find that our proposed method outperforms the state of art methods. 

\section{Introduction}

Video retrieval has been an important topic for decades. Nowadays, with the rapid development of mobile internet, more and more videos are produced and shared by users, which allows the method to retrieve videos faster and more accurate. With different query modalities, there are many types of video retrieval methods. 

The most traditional query modality is text modality \cite{7_xu2015, 8_Mithun2018, 9_youngjae2018}. At first, people focus on text-based video retrieval, which uses a keyword or a description sentence as query to search videos. However, user tagging may not be always that accurate and the expressive power is also restricted by people's description, which varies a lot among people themselves. Thus, to reduce the uncertainty of semantic expression the and to enrich the query modality's expressive power, image-based video retrieval, which uses an image as query to retrieve relevant videos, is proposed \cite{1_Yan2015Face, 2_Araujo2018Large, 3_Docampo2018}. To further enrich the expressive power, recently, video-based video retrieval, which uses another video as query, is also attractive to researchers \cite{10_Ye2013, 11_Song2018, 12_Chen2018, 13_zilos2017, 14_Zilos2019}.

However, in practical applications, videos are usually long and untrimmed, which implies that the video contain many complex actions, but only a few of them directly meet the need of users’ queries. As a result of that, a new kind of video query task called Video Moment Retrieval (VMR) is proposed, which allows user to search for certain clips inside the video instead of the whole video. 

Like the development of video retrieval mentioned above, video moment retrieval's methods are also text-based methods at first, which aims to search the video clip that is relevant to the given text query \cite{15_Hendricks2017,16_Gao2017TALL,17_Mithun2019}. However, using text as query modality still limits the richness and complexity of the information contained in the query. Then, in order to compensate for the disadvantage of text query, video-query based video moment retrieval (video relocalization) is proposed, which is also known as video relocalization task \cite{18_video_reloc,19_spatial_tempoal_video_reloc}. Its aim is temporally localizing video segments in a long and untrimmed reference video, and the segment should have semantic correspondence with the given query video clip. An example of video relocalization task is shown in Figure 1. In this example, users first pick out the clip with the action of “basketball dunk” from the query video, which is the input of video relocalization task. And the task aims at retrieving a clip with the same semantic meaning in another untrimmed reference video. 

\begin{figure*}[hbt]
	\label{fig:1 vqvmr_example}
	\centering
	\includegraphics[scale=.25]{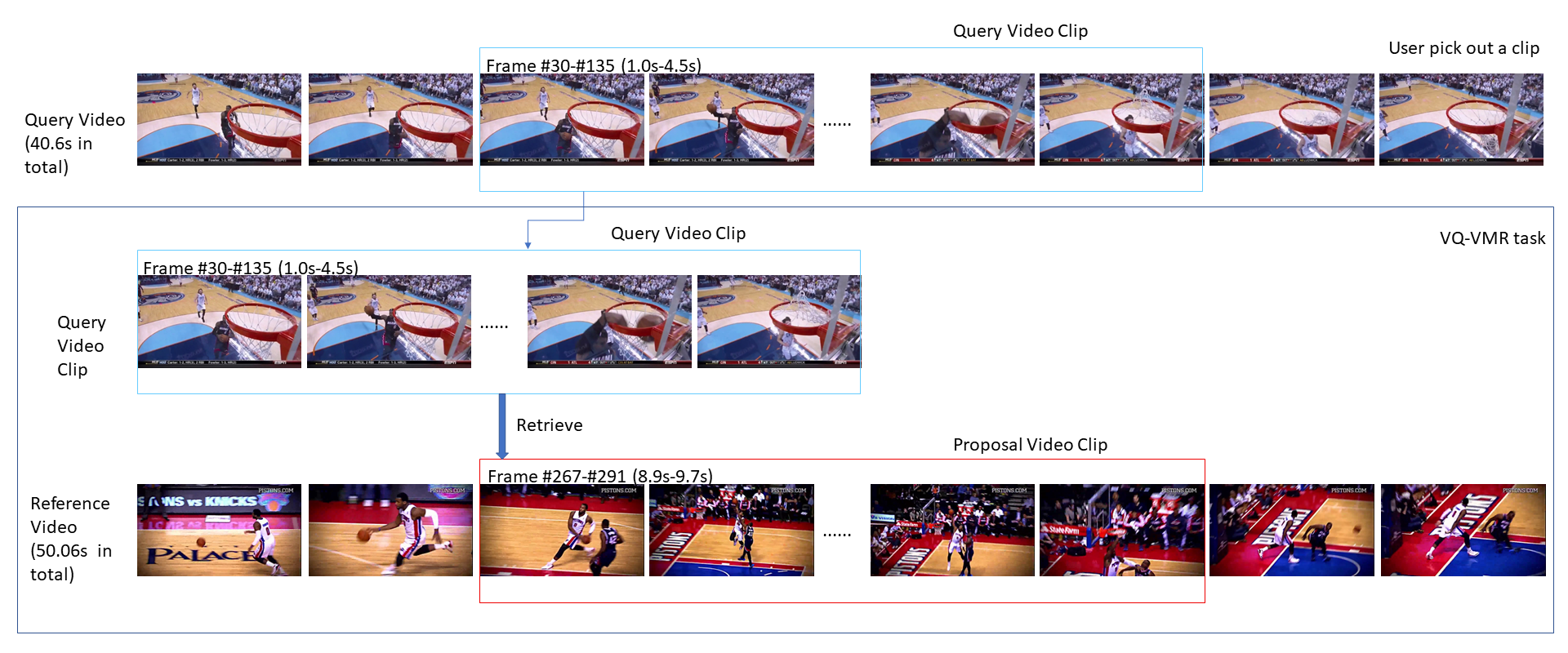}
	\caption{
		An illustration of Video Relocalization task: given a query video clip from a query video and an untrimmed  reference video, the task is to detect the starting point and the ending point of a segment in the reference video, which is semantically correspond to the query video. In this example, given a query video clip corresponding to “basketball dunk” (in blue box), video relocalization task aims to find out a clip which is also relevant to “basketball dunk” in another untrimmed long reference video (in red box). 
	}
\end{figure*}

For video relocalization task, the most direct way is leveraging semantic similarity between query video clip and reference video. \cite{18_video_reloc} made the first attempt to address this problem by proposing a cross-gated bilinear matching module. In their method, every timestep in the reference video is matched against all the timesteps in query video clip and the prediction of the starting and ending time can be regarded as a sequence labeling problem. \cite{ruolin} modified their feature extraction method by leveraging Attention-based Fusion Module to compute frame-level semantic similarity between query video clip and pre-extracted proposal clips in reference video. Then the generated Attention-based Fusion Tensor passes through Semantic Relevance Measurement Module to achieve the video-level semantic relevance between them. 

However, after observing the existing video relocalization dataset and methods, we found that there might not exists consistent relationship between feature similarity by the frame with same timestep in the two videos and feature similarity by the two videos overall, which affects fusing frame features and test result. 

Let's take the case shown in Figure 2 as an example. In the following video pairs, the two videos have the same semantic meaning (the semantic meaning is “rectification”), while the order and time of corresponding actions in the two videos vary a lot. Notice that in two videos, even for the same semantic meaning, the details are not the same. The video clip above has “adding ice” action after “cleaning the table, while the video clip below has “adding ice” at the very beginning of the clip. What’s more, some detailed sub-action appears in one-video does not appear in another video. Finally, the length of two videos are not the same, which implies the speed of the action are not the same. Thus, feature of two videos in the same timestep are not similar, although the two videos have the same semantic label and similar video-level feature. 

\begin{figure*}[hbt]
	\label{fig:2 non_corresponding_example}
	\centering
	\includegraphics[scale=.3]{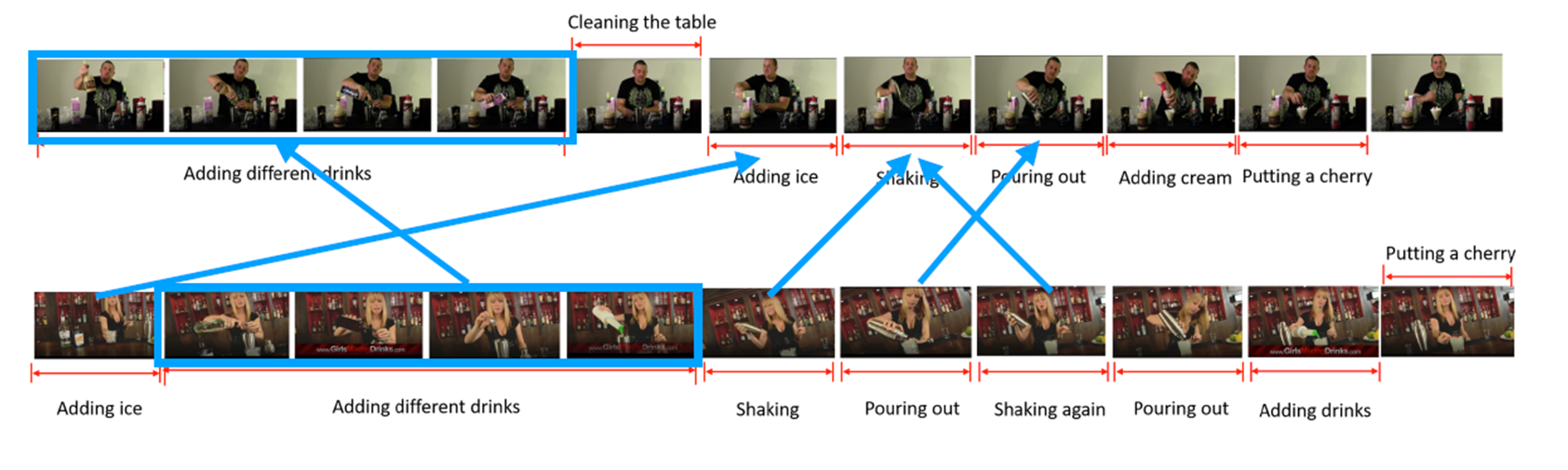}
	\caption{
	An example of the phenomenon. The 2 videos have the same semantic meaning (the semantic meaning is “rectification”). But the frame-level relationship between the 2 videos is not steady. The inconsistency does not only reflect on the visual dissimilarity, but also reflects of the details of actions. 
	Notice that in 2 videos, even for the same, the details are not the same. The video clip above has “adding ice” action after “cleaning the table, while the video clip below has “adding ice” at the very beginning of the clip. What’s more, some detailed sub-action appears in one-video does not appear in another video. Finally, the length of 2 videos are not the same, which implies the speed of the action are not the same. 
	}
\end{figure*}

One solution of it is extending feature similarity by frame to a little complicated version, where both intra-video frame similarity and inter-video frame similarity are considered to enrich feature of each frame. 

However, the former methods do not consider intra-video and inter-video frame similarity simultaneously. \cite{18_video_reloc}’s cross gated bilinear matching does not consider fusing intra-video’s features in feature fusion stage. \cite{ruolin} builds attention-based fusion tensor (AFT), which considers relationship between query video clip and proposal video. But it only uses convolutional neural network to fuse features in AFT, which only uses the local inter-video frame similarity, the global inter-video frame similarity is not used here. 

In our article, we first generate action proposals in reference video and treat query video clip and proposal clip overall as a graph, where each timestep in both video clips is treated as a node, so the feature of each timestep is the feature of each node. To fuse the features of the graph, we propose a new module called Multi-Graph Feature Fusion Module. This module leverages graph neural network to fuse features from different videos at different timesteps. What’s more, we use multiple graphs to build features of nodes at different receptive fields. Compared with the methods mentioned above, each node’s connection with nodes in two videos, thus considering both intra-video and inter-video relationship, which is very important in feature fusion stage in video relocalization task. 

Our main contributions of this paper can be summarized as follows:

1. We found a phenomenon on non-consistent feature relationship between frame-level and video-level, thus affecting feature fusion and test result of our video relocalization task. We try to alleviate this by treating query video clip and proposal overall as a graph. 

2. We propose a new module called Multi-Graph Feature Fusion module to fuse nodes’ features of the graph proposed above. 

3. We conduct extensive experiments on ActivityNet v1.2 dataset and Thumos14 dataset, and the results show that our method outperforms many state-of-the-art methods. 

4. We conduct ablation study of our proposed method, proving its effectiveness and finding the architecture with best result.

\section{Related Work}

\subsection{Video Retrieval}

Video retrieval aims at selecting the video which is most relevant to the query video clip from a set of candidate videos. According to different types of query modalities, video retrieval can be divided into following categories: text-query based video retrieval, image-query based video retrieval and video-query based video retrieval. 

Text-query based video retrieval has long been tackled via joint visual-language embedding models \cite{23_Torabi2016,24_KirosSZ14,25_Hodosh2013Framing,26_Lin2014,27_Vendrov2015,28_Hu2016,29_Mao2016}. Recently, much progress has been made in this aspect. Although text and video are different modalities, which brought difficulties in studying joint feature representations, some earlier works still manage to achieve good results. Several deep video-text embedding methods \cite{23_Torabi2016,30_Xu2015Jointly,31_Otani2016Learning} has been developed by extending image-text embeddings \cite{32_Devise,33_Socher2013Grounded}. Other recent methods improve the results by utilizing concept words as semantic priors \cite{34_Yu2017}, or relying on strong representation of videos such as Recurrent Neural Network-Fisher Vector (RNN-FV) method \cite{36_KaufmanLHW16}. Also, some dominant approach leverages RNN or their variants to encode the whole multimodal sequences (e.g. \cite{23_Torabi2016,34_Yu2017,36_KaufmanLHW16}). which faces the challenge of processing cross-modality data. However, text and videos are different modalities, which means there exists inconsistency between features from the two modalities. 

Image-query based video retrieval techniques uses an image as query. Li et.al \cite{1_Yan2015Face} propose Hashing across Euclidean space and Riemannian manifold (HER) to deal with the problem of retrieving videos of a specific person given his/her face image as query. A. Araujo et.al \cite{2_Araujo2018Large} introduce a new retrieval architecture, in which the image query can be compared directly with database videos. Although compared with text modality’s feature, image modality’s feature is much more like video modality’s feature, image modality’s feature only provide appearance at one certain time, thus lacks dynamic clues.

As the expression power of text and single image are always limited, video-query based retrieval techniques are proposed to break this kind of limitation. Some video-based methods still borrow the idea of hashing from image-query based video retrieval which also map high-dimensional video features to compact binary codes so that it can address video-to-video retrieval techniques. And video retrieval has many specific applications, such as fine-grained incident video retrieval and near-duplicate video retrieval, which mainly focus on retrieving videos of the same incident and duplicated videos respectively \cite{13_zilos2017,14_Zilos2019,37_Zilos2019}.

However, in practice, videos are still very long and untrimmed. But only the clip with certain action directly meets the user’s need. To this end, video moment retrieval task is proposed, which only retrieves the video clip with certain action given a query. And our paper focuses on this task. 

\subsection{Temporal Proposal Generation}

Temporal Proposal Generation is used in Temporal Action Localization task to generate proposals of actions. Earlier proposal generation method is sliding window method, which slides the temporal window along time dimension to pick out candidates. Based on sliding windows method, \cite{38_Gaidon2013,39_Yuan2016,40_Jain2014,41_Ma2013,42_Heilbron2016} uses proposal network to predict if the current sliding window is action or not, so that some sliding windows are removed. However, sliding window method is not always satisfying, for the length of the sequence is fixed, while different actions last different time. To solve the problem, Heilbron et. al. \cite{fast_Heilbron2016Fast} propose Fast Temporal Activity Proposals method. Escorcia et. al. \cite{dap} proposes Deep Action Proposals (DAP) method to generate proposals, in which a visual encoder, a sequence encoder, a localization module and a prediction module are composited as a pipeline to extract K proposals with confidence scores over a T timestep video. Zhao et. al. \cite{49_tag_proposal} proposed a method called Temporal Actionness Grouping method. They use an actionness classifier to evaluate the binary actionness probabilities for individual snippets and a repurposed watershed algorithm to combine the snippets as proposals. In our article, we need temporal proposal generation method to generate raw proposal clip in query videos and their reference videos, and we use TAG method in \cite{49_tag_proposal} to generate our proposals. 

\subsection{Video Moment Retrieval}

Derived from video retrieval, Video Moment Retrieval needs to find out semantic-relevant clips in a video given a query. It can also be divided into two main research methods: text-query based video moment retrieval and video-query based video moment retrieval, and the latter one is also called "video relocalization" \cite{18_video_reloc}. 

Text-query based video moment retrieval focus on locating temporal segment which is the most relevant to the given text. Hendricks et. al. \cite{15_Hendricks2017} propose Moment Context Network which leverages both local and global video features over video's timesteps and effectively realize the localization in videos based on natural language queries. Gao et. al. \cite{16_Gao2017TALL} propose a Cross-Modal Temporal Regression Localizer to jointly model both textual query and video candidate moments, and its localizer outputs alignment scores between them and action's regression boundaries. With the development of attention mechanism in the field of vision and language interaction, attention is gradually used in Video Moment Retrieval models to help capturing interactions between text and video modalities. Both matching score and boundary regression are also considered in our work. We put these thoughts in our work to make it reasonable. 

Different from text-query based video moment retrieval, video-query based video moment retrieval does not have the cross-modality problem, since both query and reference are both from video modality. The methods of this part are very few. \cite{18_video_reloc} make the first attempt on by using a cross-gated bilinear matching module. In this method, the feature in reference video at every timestep is matched with every timestep in query video clip via attention mechanism. Thus, based on matching  Later on, \cite{ruolin} improved the result by using Attention-based Fusion module and Semantic Relevance Measurement module to capture frame-level relationship, however, this method still treats video relocalization task as a traditional regression problem. \cite{19_spatial_tempoal_video_reloc} extends this task to spatial-temporal level, which requires to find out both temporal segment and spatial location in the proposal video given a query video clip. In our article, our task is just video moment retrieval, which is the same as the task in \cite{18_video_reloc} and \cite{ruolin}. 

\subsection{Graph Neural Networks}

Graph Neural Networks were firstly proposed in \cite{gnn_first}, which are used for node classification, graph classification and link prediction tasks at first. 

With the success of Graph Neural Network in many aforementioned graph tasks, Graph Neural Network shows it strong power in extracting features of graph data. And many other non-graph tasks also begin using graph neural networks: they first model the input data of their tasks as graphs, and then use graph neural network to extract and fuse features. For example, \cite{gnn_image} uses graph neural networks in Image Denoising task, where a pixel is treated as a node in the graph. \cite{gnn_video} uses graph neural network in video semantic segmentation task, where each timestep is treated as a node in the graph. 

After \cite{gnn_first}, many new kinds of Graph Neural Networks are proposed, and they can be divided into two aspects: spectral based methods and spatial based methods. 

Spectral based focus on interpreting Graph Neural Networks from graph spectrums and Graph Fourier Transforms. And Laplacian matrix (which represents the graph spectrum) is calculated in this kind of methods. Graph Convolutional Network (GCN) \cite{gcn} and Graph Attention Network (GAT) \cite{gat} are two examples. 

However, spatial based methods focus on message passing from current node’s neighbors to current node, and nodes’ features are updated directly via their neighbors (and no graph spectrum information is used). GraphSAGE \cite{graphsage} is a typical example. 

For this task, we treat correlations among different timesteps as a graph data, which better represents the frame-level relationship, and we use Graph Neural Network method to fuse the feature of all timesteps in two videos. This graph modeling scheme is the same as that in \cite{gnn_video}. 

As for graph neural network methods, our method is more likely to be a spatial based method than a spectral one, for we just use the original connections among nodes in our defined graphs, and we do not utilize the spectral information of those graphs.

\section{Our Proposed Method}

In this section, we will introduce our proposed method for Video Query Based Video Retrieval task, which aims at finding the most semantic relevant clip in a long untrimmed video given a query video clip.

\subsection{Problem Formulation}

Given an query video clip Q, and a reference long video P. Our target is to get starting point and ending point $[s_{pred}, e_{pred}]$ of video clip inside P. 

To achieve this goal, in training stage, we use triplet $(q,p,n)$ as input, where $(q,p,n)$ denote query video clip, positive video (same semantic label with query) and negative video (different semantic label with query) respectively, and the total architecture of our purposed method is shown in Figure 3. This method is different from that in \cite{18_video_reloc}, where video-query based method is treated as a sequence labeling problem. We use triplets and triplet loss to train the model so that it can better feature fusion, and regression loss is also used for refining positive sample's $[s_{pred}^{train}, e_{pred}^{train}]$ in training stage.

\begin{figure*}[hbt]
	\label{fig:3 our proposed method}
	\centering
	\includegraphics[scale=.3]{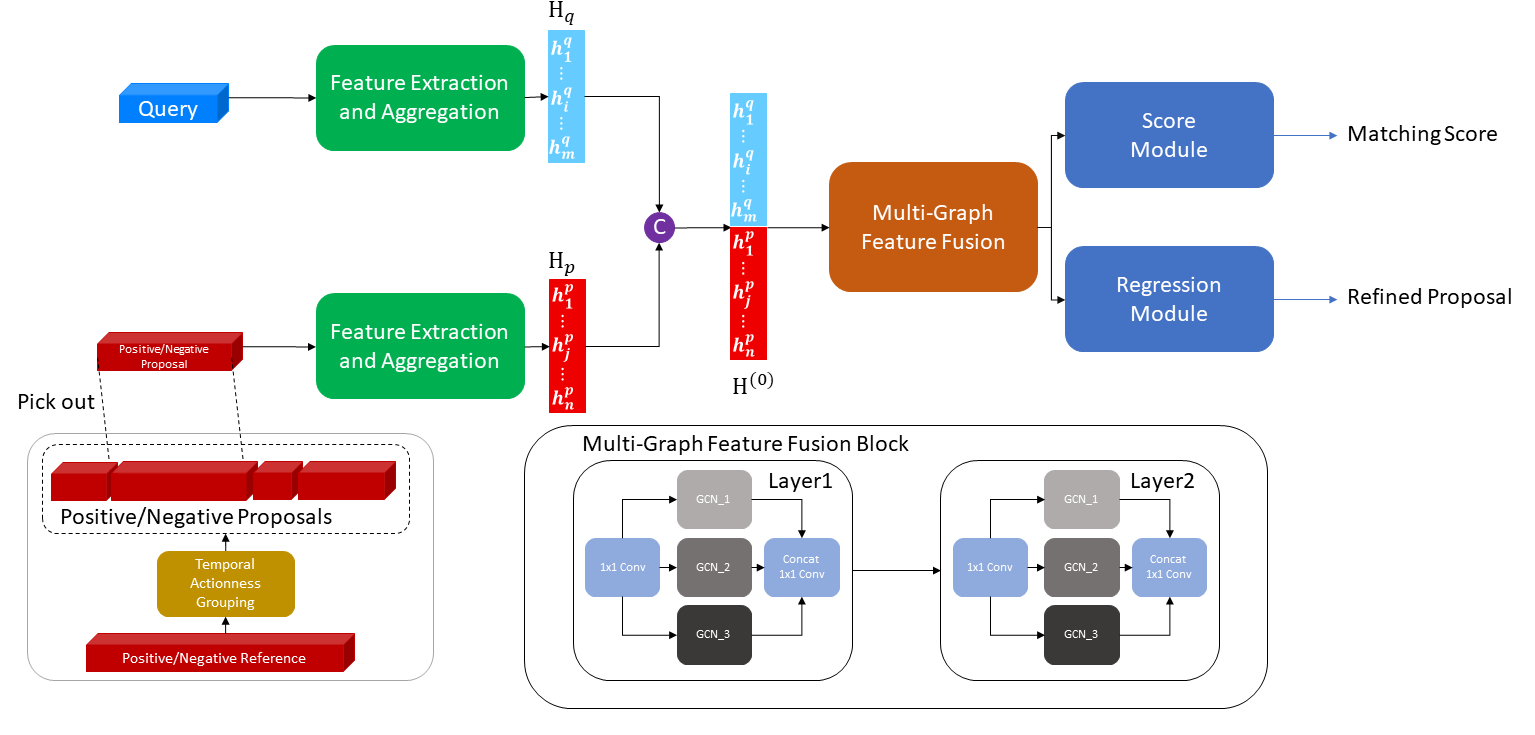}
	\caption{
	The total architecture of our module. The key architecture is Multi-Graph Feature Fusion Module, which leverages graph convolution to extract the features from different videos at different timesteps. In Multi-Graph Feature Fusion Module, we use multiple graph convolution layers to extract features at different scales. 
}
\end{figure*}

The training stage can be summarized as follows: first, input positive/negative video is passed to TAG backbone \cite{49_tag_proposal} to get corresponding proposals. Among all the proposals, we pick one of the proposals out to represent the positive/negative video. Then video feature of query video clip and each proposal are fed into LSTM Module \cite{lstm} to extract global feature matrices. After concatenating query clip’s feature matrix and proposal’s feature matrix into one matrix along time dimension, the concatenated matrix is fed into Multi-Graph Feature Fusion Module to fuse intra-video and inter-video features. Finally, the features are sent to Score Module and Regression Module to calculate triplet loss and regression loss. 

In testing stage, we do not use negative sample of the triplet, and only $(q,p)$ pairs are used in testing stage. For one query video clip, we also use TAG method to pick out all the proposals in the positive video. Different from picking one proposal out in training stage, we use all the proposals to predict each proposal's $[s_{pred}^{test} e_{pred}^{test}]$ as output. 

In the following paragraph, we will describe our method in detail: section 3.1 is about video feature encoding, 3.2 is about building weighted adjacency graphs, 3.3 is about Multi-Graph Feature Fusion Block, and 3.4 is about Score Module and Regression Module. Since positive and negative samples are passed into the same branch in our proposed method, and in the following paragraph, if not specified, we only take positive sample as example. 

\subsection{Video Feature Encoding via LSTM}

For both query video clip and positive/negative proposal video clip (in the following paragraph, if not specified, we use proposal to denote positive/negative proposal video clip), we use C3D Module \cite{50_C3D} to extract video's feature. The original output dimension of C3D Module is 4096, and we use Principle Component Analysis (PCA) to project the feature vector to a 500-dimensional feature space to reduce input dimensions. Since C3D Module only consider short range feature characteristics, we use two Long Short Term Memory (LSTM) units for query and proposal features to aggregate the extracted features in a longer time range:

$$h_{q,t}=\mathrm{LSTM}(q_t, h_{q, t-1}) \quad t=0,1,...,m-1$$
$$h_{p,t}=\mathrm{LSTM}(p_t, h_{p, t-1}) \quad t=0,1,...,n-1$$

Where $q_t,p_t \in \mathbb{R}^d$ denotes input query and proposal video feature at step t respectively. $h_{q,t},h_{p,t} \in \mathbb{R}^d$ denotes hidden query feature and hidden proposal feature at step $t$ respectively. And $\mathrm{LSTM}(\cdot,\cdot)$ denotes a Long Short Term Memeory Block. 

After achieving hidden video features via two LSTM blocks, we concatenate output $h_{q,t}$ and $h_{p,t}$ along time dimension, which can be represented as follows:

$$\mathrm{H}_q = [h_{q,0};...;h_{q,i};...;h_{q,m-1}]\in\mathbb{R}^{m \times d}$$
$$\mathrm{H}_p = [h_{p,0};...;h_{p,i};...;h_{p,n-1}]\in\mathbb{R}^{n \times d}$$

Where $[...;...;...]$ denotes concatenate along time dimension. 
Then, in order to fully utilize the relationship between query part and proposal part, we concatenate the two feature matrices $\mathrm{H}_q$ and $\mathrm{H}_p$ along time dimension to form a feature matrix $H^{(0)}$, which is ready for the input of Multi-Graph Feature Fusion Module:

$$\mathrm{H}^{(0)}=[ \mathrm{H}_q; \mathrm{H}_p] \in \mathbb{R}^{(m+n) \times d}$$

In our proposed method, we set $m=n=T=40$, so we have $\mathrm{H}^{(0)} \in \mathbb{R}^{2T \times d}$, and this is the input of our multi-graph feature fusion block. 

\subsection{Building adjacency matrix}

After aggregating feature in LSTM module and two concatenations, we have concatenated feature matrix $\mathrm{H}^{(0)}$. For $\mathrm{H}^{(0)} \in \mathbb{R}^{2T\times d}$, we treat $2T$ rows as $2T$ nodes in the graph. So, each row’s feature is treated as each node’s feature. As described above, we get the graph’s feature matrix $\mathrm{H}^{(0)}$, which is an input in graph convolution. In the next section, we will talk about building edges, i.e., building binary weighted adjacency matrix $\mathrm{\hat{A}}_k$, where $k$ is a parameter related to different kinds of graphs. Since we have $2T$ nodes, the adjacency matrix $\mathrm{\hat{A}}_k$ is a $2T \times 2T$ matrix. 
Since the first $T$ nodes are from query video clip, while the last $T$ nodes are from proposal video, we divide $\mathrm{\hat{A}}_k$ into 4 submatrices: $\mathrm{\hat{A}}_{qq,k},\mathrm{\hat{A}}_{qp,k},\mathrm{\hat{A}}_{pq,k},\mathrm{\hat{A}}_{pp,k} \in \mathbb{R}^{T \times T}$ as shown in Figure 4. 

\begin{figure*}[hbt]
	\label{fig:4 about adjacency}
	\centering
	\includegraphics[scale=.3]{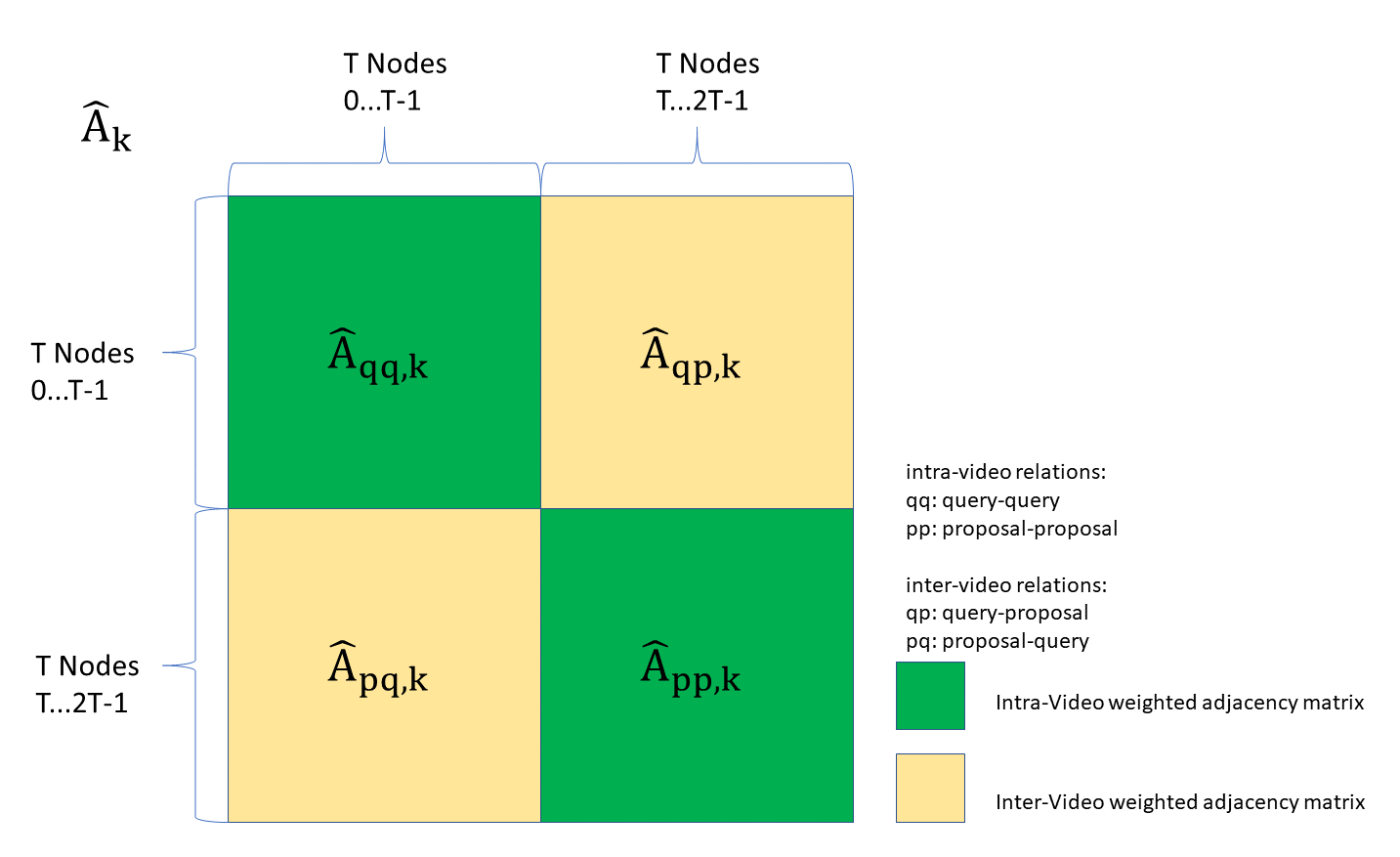}
	\caption{
		The weighted adjacency matrix $\mathrm{\hat{A}}_k$ is divided into four submatrices, $\mathrm{\hat{A}}_{qq,k},\mathrm{\hat{A}}_{pp,k} \in \mathbb{R}^{T \times T}$) (in yellow) reflects intra-video relationship, while$\mathrm{\hat{A}}_{qp,k},\mathrm{\hat{A}}_{pq,k} \in \mathbb{R}^{T \times T}$ (in green) reflects inter-video relationship. 
	}
\end{figure*}

It is clear to show that $\mathrm{\hat{A}}_{qq,k}$ and $\mathrm{\hat{A}}_{pp,k}$ reflects intra-video node connections, while $\mathrm{\hat{A}}_{qp,k}$ and $\mathrm{\hat{A}}_{pq,k}$ reflects inter-video node connections. In the following part, we will talk about our methods of building inter-video node connections and intra-video node connections. 

\subsubsection{Intra-Video node connections}

First, we consider inter-video node connections, i.e., node connections inside query/proposal video. The cases in query video clip and proposal video are the same, so we have $\mathrm{\hat{A}}_{pp,k} = \mathrm{\hat{A}}_{qq,k}$. 

As a result, we just consider building $\mathrm{\hat{A}}_{qq,k}$ in the following words. For each node $v_i, i=0,1,…T-1$, we connect the node with its previous $k$ nodes and next $k$ nodes in the same video (connect the node with the nodes in a range of $k$). As to boundary nodes of a video (for example, node $v_0$ and node $v_{T-1}$ in query video clip sequence), we ignore the missing connection nodes of them, i.e., their intra-video node connections are less than $2k$. The intuition of setting this kind of connection is that intra-video node connections should only consider connections in a small range to reduce interactions, for one video itself is consecutive in time. On the other side, if we use fully connected weighted adjacency submatrix $\mathrm{\hat{A}}_{qq}$ and $\mathrm{\hat{A}}_{pp}$ to build the graph, they have too many weighted connections, which may have negative effect on the result.

In all, the described above can be written as follows: 

$$\mathrm{\hat{A}}_k[i,\max(0,i-k):\min(T-1,i+k)+1:1]=1.0$$

where $start:stop:step$ implies $[start,start+step,start+2 \times step...,stop)$ sequence. Note that $\max$ and $\min$ in this formula limit the range of the video sequence. In all, with this method, we can have symmetric adjacency matrices $\mathrm{\hat{A}}_{pp,k}$ and $\mathrm{\hat{A}}_{qq,k}$. 

\subsubsection{Inter-Video node connections}

After building intra-video node connections, we consider building the rest part of the weighted adjacency matrix, i.e. $\mathrm{\hat{A}}_{qp, k}$ and $\mathrm{\hat{A}}_{pq, k}$, which represents inter-video node connections. 

Notice that the weighted adjacency matrix $\mathrm{\hat{A}}_k$ is symmetric, so $\mathrm{\hat{A}}_{pq,k}=\mathrm{\hat{A}}_{qp,k}^{\top}$, and it is enough for us to consider building $\mathrm{\hat{A}}_{qp,k}$ only.

To build $\mathrm{\hat{A}}_{qp,k}$, we connect node in query video clip with nodes in proposal video in the same timestep and nodes on both sides with a stride k. It can be written as follows:

$$\mathrm{\hat{A}}_{k}[i,T+i:T-1:-k]=1.0$$
$$\mathrm{\hat{A}}_{k}[i,T+i:21:k]=1.0$$

where $start:stop:step$’s definition is the same as that shown in Intra-Video node connections. 

The intention of building graphs is that there might not exists consistent relationship between feature similarity by the frame with same timestep in the two videos and feature similarity by the two videos overall, in order to make our fused frame features more consistent with the video’s overall feature, we need to consider local intra-video timestep connections and global inter-video timestep connections. We build such connections as graphs for better fusing features. 

To show this procedure much more intuitively, Figure 5  shows the connection. And an example of building such a graph is also illustrated in Figure 6. In our proposed method, we use multiple k (for example, k=1,2,3) to build multiple graphs. With these graphs, more relations among nodes can be expressed, so that we can extract features of multiple scales.

\begin{figure*}[hbt]
	\label{fig:5 adjacency building rules}
	\centering
	\includegraphics[scale=.3]{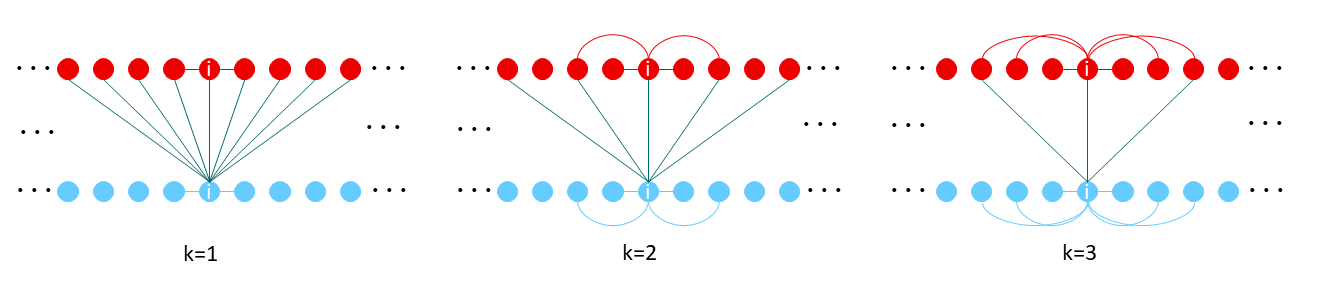}
	\caption{
		The connection of current node $v_i$ in different types of weight adjacency matrices $\mathrm{\hat{A}}_k$s. The blue nodes denote timesteps in query video, and the red nodes denote timesteps in proposal video. The blue edges and red edges denote intra-video node connections, while the green edges denote inter-video node connections. 
	}
\end{figure*}

\begin{figure*}[hbt]
	\label{fig:6 an example of a graph}
	\centering
	\includegraphics[scale=.3]{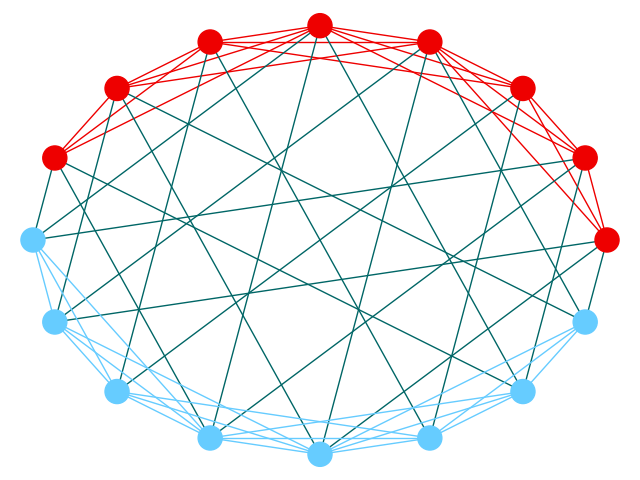}
	\caption{
		An example of our graph building method. In this example, T=8 (8 blue nodes as query and 8 red nodes as proposal), and k=3 (see the building method for detail)
	}
\end{figure*}

\subsection{Graph Convolution} 

After building the weighted adjacency matrix $\mathrm{\hat{A}}_k$, which reflects relation between query video clips and proposals, we can use graph convolution to fuse features of query and proposal videos. In this part, we will describe the graph convolution in our proposed method.
A graph convolution operation can be written as follows:

$$ \mathrm{H}^{(l+1)}=\sigma(\mathrm{\hat{A}}\mathrm{H}^{(l)}\mathrm{W}) $$

where $\mathrm{H}^{(l)} \in \mathbb{R}^{N \times F_{in}}$ denotes feature matrix in the $l^{th}$ layer, $\mathrm{\hat{A}} \in \mathbb{R}^{N \times N}$ denotes weighted adjacency matrix, $\mathrm{W} \in \mathbb{R}^{F_{in} \times F_{out}}$ denotes trainable weight matrix. $\sigma(\cdot)$ denotes nonlinearity function.
There are many types of weights for $\mathrm{\hat{A}}$, for example:

1. Binary weights. This is the original definition of adjacency matrix. 
2. Exponential weights.
$\mathrm{\hat{A}}[i,j]=\exp(-\frac{\vert i-j \vert^2}{2s^2}) \qquad i,j\in\{0,1,...,N-1\}$. where $i$ and $j$ are nodes' indicies, and $s$ is a hyperparamter. 

3. Laplacian weights. 
$\mathrm{\hat{A}}=\tilde{D}^{-0.5}\tilde{A}\tilde{D}^{-0.5}$. Where $\tilde{A}=\mathrm{A}+I_N$ is an adjacency matrix with self-loop, $\tilde{D}=\mathrm{diag}(\sum_j{\mathrm{\tilde{A}[i,j]}})$ is the diagonal degree matrix of $\mathrm{\tilde{A}}$. 

As described in section 3.2, we use the first type of weights, i.e., binary weighted adjacency matrix, for we just want to propagate the feature among the frames with the rule of natural connection, and we do not care much about the relationship with weight and other kinds of information in this task.

\subsection{Multi-Graph Feature Fusion} 

After building the proper adjacency matrix, we have all the necessities of a graph convolution layer. Then we use $\mathrm{H}^{(0)}$ and $\mathrm{\hat{A}}_k$ as the inputs of our Multi-Graph Feature Fusion Layer, where graph convolution is used inside. The detailed architecture of Multi-Graph Feature Fusion Layer can be seen in Figure 7. 

In every Multi-Graph Feature Fusion Layer, we first use an $1\times1$ convolution to project features to a latent feature space. Then we use graph convolutions mentioned above, with different $\mathrm{\hat{A}}_k$ as adjacency matrix, i.e. $\mathrm{\hat{A}}_1,\mathrm{\hat{A}}_2$, to extract multi-scale intra-video and inter-video correlations. After graph convolution with multiple graphs, all these features are concatenated along feature dimension, and a $1\times1$ convolution is used to fuse features of multiple scales. 

In all, the whole procedure of Multi-Graph Feature Fusion Layer can be written as:

$$\mathrm{\tilde{H}}^{(l)}=\mathrm{Conv_{1\times1}}(\mathrm{H}^{(l)})$$

$$\mathrm{H}^{(l+1)}=\mathrm{Conv_{1\times1}}([\sigma(\mathrm{\hat{{A}}}_1 \mathrm{\tilde{H}}^{(l)} \mathrm{W}_1^{(l)}),..., \sigma(\mathrm{\hat{{A}}}_k \mathrm{\tilde{H}}^{(l)} \mathrm{W}_k^{(l)})])$$

Where $\mathrm{Conv}_{1\times1}(\cdot)$ denotes  convolution, $[...,...,...]$ denotes concatenate along feature dimension. $\mathrm{\tilde{H}}^{(l)} \in \mathbb{R}^{2T\times d^{(l)}}$ is the result after the first $1\times1$ convolution. $\mathrm{W}_{k}^{(l)} \in \mathbb{R}^{d^{(l)}\times d^{(l+1)}}$ is trainable weight matrix in graph convolution, $\sigma(\cdot)$ is the nonlinearity function. In our experiments, we use $\tanh(\cdot)$ as nonlinearity function. 

We use $L$ cascaded Multi-Graph Feature Fusion layers to build this module, so that we can make full use of node connections inside a video and among videos.
\begin{figure*}[hbt]
	
	\label{fig:7 our architecture of Multi-Graph Feature Fusion}
	\centering
	\includegraphics[scale=.3]{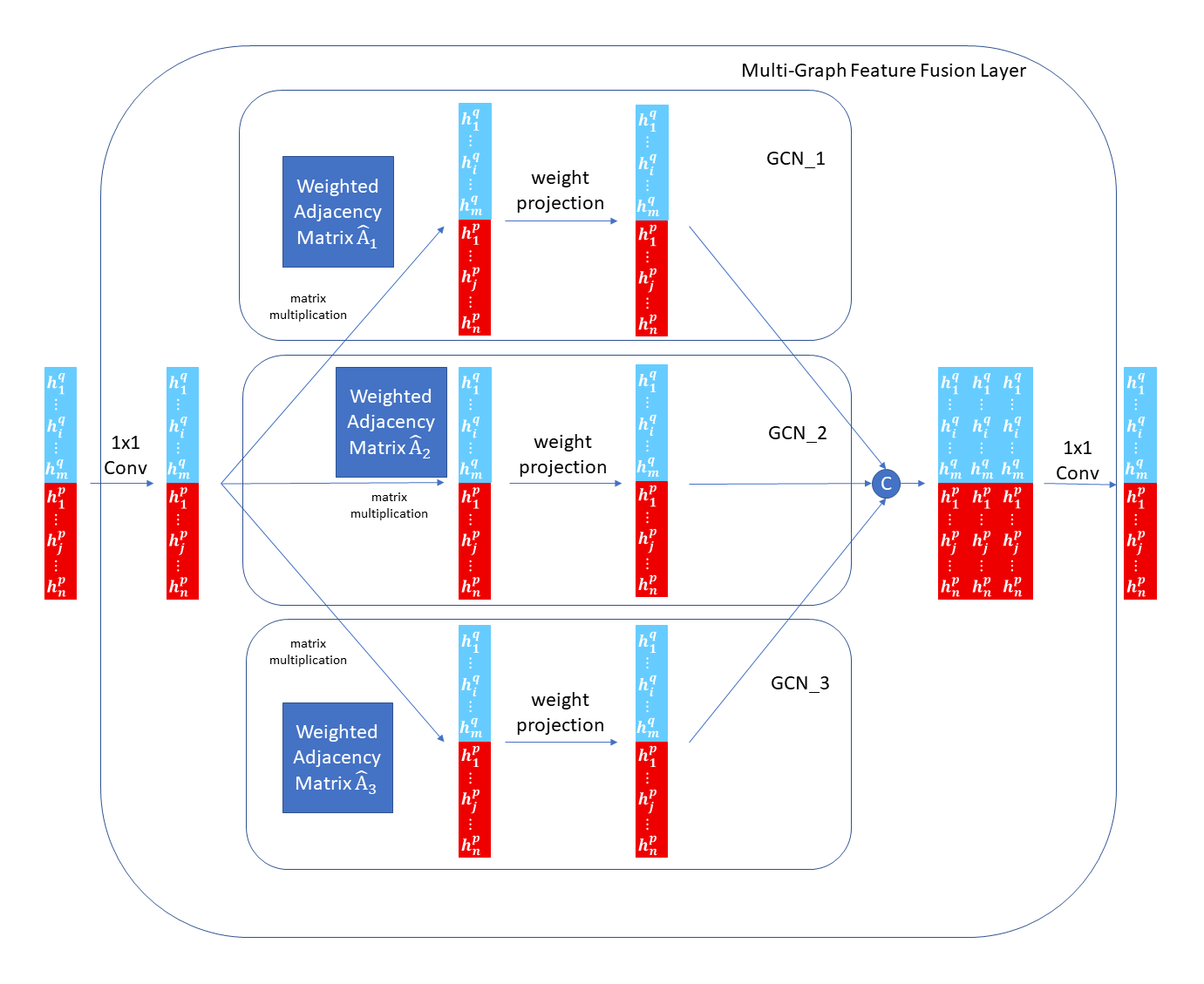}
	\caption{
		The detailed architecture of one Multi-Graph Feature Fusion Layer. First, the concatenated video feature pass through an $1\times1$ convolution block, then, with different $\mathrm{\hat{A}}_k$ as graph adjacency matrices, the graph feature passes through different graph convolution blocks. Then the features extracted from different graph convolution blocks are concatenated along time dimension and finally, an $1\times1$ convolution is applied to further fuse those features. 
	}
\end{figure*}

\subsection{Score module and Regression Module}

After fusing the via Multi-Graph Feature Fusion Module, we use global average pooling to the feature of 2T nodes $\mathrm{H}^{(L)}$ into one node’s feature.

$$h_{\mathrm{glboal}}=\mathrm{AvgPool}(\mathrm{H}^{(L)})\in\mathbb{R}^{1\times d^{(L)}}$$

where $h_{global}$ denotes output global feature, $\mathrm{AvgPool}(\cdot)$ denotes average pooling, $d^{(L)}$ denotes output feature of final Multi-Graph Feature fusion layer. Then $h_{global}$ is fed into score module and regression module. 

\subsubsection{Score Module}

$h_{global}$ is fed into this module, which consists a Multi-Layer Perception (MLP), and the module outputs $s\in [-1,1]$, which represents the similarity score between two videos. And the output $s$ is used for Triplet Loss. 

\subsubsection{Regression Module}

$h_{global}$ is fed into this module, which also consists an MLP, and the module outputs regression offsets $[T_c, T_l]\in \mathbb{R}^2$, where $T_c$ and $T_l$ are related to central point offset and length offset. Since the proposal clip can be either tight or loose, this regression progress tends to refine the proposal extracted from TAG module. 

\subsection{Losses}

Our total loss function is made up of two items: triplet loss is used for extracting and fusing features of between query and proposals, and regression loss is used for refining starting and ending points. 

The loss funciton is defined as follows: 

$$L_{total} = L_{tri} + \mu L_{reg}$$

where $L_{tri}$ denotes triplet loss and $L_{reg}$ denotes regression loss. $\mu$ is a weight balancing term. In our experiments, $\mu$ is set to be 1.0. 

\subsubsection{Triplet Loss}

The training samples are in the form of $(q, p, n)$, where $(q, p, n)$ denotes query video clip, positive video, negative video respectively. And the triplet loss comprises 2 kinds of similarity scores: positive-query similarity score and negative-query similarity score. $L_{tri}$ can be written as follows: 

$$L_{tri}=\sum_{i=1}^{N}\max(0,\gamma-S(q,p)+S(q,n))+\lambda\Vert \theta \Vert_2^2$$

where $N$ is batch size. $\gamma$ is a hyper parameter to ensure a sufficiently large difference between the positive-query score and negative-query score. $\lambda$ is a hyper parameter on regularization loss. $S(\cdot, \cdot)$ denotes similarity score between two video sequences. In our experiment, we set $\gamma=0.5$, and $\lambda=5\times10^{-3}$. 

\subsubsection{Regression Loss}

Only the positive proposals are used to train the regression loss, since negative proposal . In our experiments, regression loss $L_{reg}$ is in the form that follows:

$$L_{reg} = \frac{1}{N} \sum_{i=1}^{N} \vert T_{c,i} - T_{c,i}^* \vert + \vert T_{l,i} - T_{l,i}^* \vert$$

Where $T_{c,i}$ and $T_{l,i}$ are predicted $i^{th}$ positive proposal’s relative central point and offset. $T_{c,i}^{*}$ and $T_{l,i}^{*}$ are ground truth which are calculated as follows:

$$ T_{l,i}^{*} = \log( {\frac{\mathrm{len}_i}{\mathrm{len}_i^*}}) $$

$$ T_{c,i}^{*} = \frac{\mathrm{loc}_i-\mathrm{loc}_i^*}{\mathrm{len}_i^*} $$

Where $\mathrm{loc}_i$ and $\mathrm{len}_i$ denote the center coordinate and length of the $i^{th}$ proposal respectively, $\mathrm{loc}_{i}^{*}$ and $\mathrm{len}_{i}^{*}$ denote those of the corresponding ground truth segments.

\subsection{Testing Stage}

After training our proposed framework, we perform the task on test set. Testing stage aims at retrieving the matching clip in an untrimmed video given a query clip. As a result, we only use the reference video which is known to have the same semantic label as query video clip, and no "negative" video is used in test stage. The query video clip and reference video mentioned above are paired to be our input in test stage. 

There are two procedures in testing stage: proposal selection and proposal refinement. 

\subsubsection{Proposal selection}

Given a query video clip $q$, we first get M proposals of the reference video using TAG method. Then, we calculate the score between query video clip and each of the M proposals. And the proposal with highest score is selected, which can be expressed as

$$m=\arg \max_m S(q, p_m)$$

\subsubsection{Proposal Refinement} 

After selecting the proposal with highest score, whose index is $m$, the boundary of the $m^{th}$ proposal is then refined based on the regression module in Figure 3. The refinement procedure is described as follows. 

In Regression Module, we have:

$$T_{l,i,m}^*=\log ( {\frac{\mathrm{len}_{i,m}}{\mathrm{len}_{i,m}^*}})$$

$$T_{c,i,m}^*=\frac{\mathrm{loc}_{i,m}-\mathrm{loc}_{i,m}^*}{\mathrm{len}_{i,m}^*}$$

Where ${\mathrm loc}_{i,m}$ and ${\mathrm len}_{i,m}$ denote the predicted center coordinate and length in the $m^{th}$ proposal of $i^{th}$ reference video, and $\mathrm{loc}_{i,m}^{\ast}$ and $\mathrm{len}_{i,m}^\ast$ denote those of the corresponding ground truth segments. In testing stage, our target is to refine the proposal with highest score, i.e., we want to predict the "ground truth" $\mathrm{loc}_{i,m}^{\ast,pred}$ and $\mathrm{len}_{i,m}^{\ast,pred}$ with given proposal $\mathrm{loc}_{i,m}$, $\mathrm{len}_{i,m}$ and calculated $[T_{c,i,m}^\ast,\ T_{l,i,m}^\ast]\in\mathbb{R}^2$.

As a result, we have

$$\mathrm{len}_{i,m}^{*,pred}=\exp(-T_{l,i,m}^{*,test})\mathrm{len}_{i,m}$$

$$\mathrm{loc}_{i,m}^{*,pred}=\mathrm{loc}_{i,m}-T_{c,i,m}^{*,test}\mathrm{len}_{i,m}^{*,pred}=\mathrm{loc}_{i,m}-T_{c,i,m}^{*,test}\exp(-T_{l,i,m}^{*,test})\mathrm{len}_{i,m}$$

Where $T_{l,i,m}^{\ast,pred}$ and $T_{c,i,m}^{\ast,pred}$ are the calculated $T_{l,i}^\ast$ and $T_{c,i}^\ast$ in test phase respectively. 

Then we have the refined starting and ending point $s_{pred}$ and $e_{pred}$:

$$s_{pred}=\mathrm{loc}_{i,m}^{*,pred}-\frac{1}{2}\mathrm{len}_{i,m}^{*,pred}$$
$$e_{pred}=\mathrm{loc}_{i,m}^{*,pred}+\frac{1}{2}\mathrm{len}_{i,m}^{*,pred}$$

With the refined starting and ending point, we get our proposal clip in the reference video.

\section{Experiments}

In this section, we conduct experiments on both ActivityNet v1.2 dataset \cite{21_ActivityNet} and Thumos14 dataset \cite{22_THUMOS14}, and the former one is often used in video relocalization task. As the result shows below, our proposed method achieves better performance comparing with existing methods. 

This section is organized as follows: first, we introduce datasets and implementation details. Then we introduce our method’s results with other methods’ on ActivityNet v1.2 dataset and Thumos14 dataset. Finally, we introduce ablation study of our proposed method on ActivityNet Dataset. 

\subsection{Datasets}

As for video relocalization task, \cite{18_video_reloc} first exploit and reorganize the videos in ActivityNet to form a new dataset for research. Also, we added experiments on Thumos14 dataset, which is the same as \cite{ruolin}, to prove the effectiveness of our proposed method. 

In both datasets, original videos are annotated with starting and ending point for each  action, which is referred to "ground truth" in the following paragraph.

\subsubsection{ActivityNet}

ActivityNet v1.2 \cite{21_ActivityNet} has 9682 videos, which are divided into 100 action classes. We reorganized ActivityNet v1.2 for our study. Following the split methods in \cite{18_video_reloc}, we split 80 classes, 10 classes, 10 classes for training, validation, testing respectively. In the experiment, we use the pre-extracted 500-dimension PCA features with a temporal resolution of 16. 

\subsubsection{Thumos14}

Thumos14 dataset \cite{22_THUMOS14} has many videos, but the untrimmed long videos with temporal annotations directly meet our needs. We picked out 412 of them (200 from validation data and 212 from test data in the original Thumos14 dataset) for our training and testing, which are from 20 classes. We randomly select 14 classes for training and the rest 6 classes for testing. We need to remind that two falsely annotated videos ("270" and "1496") in the testing set were excluded in the present study. 

\subsection{Implementation Details}

The total training procedure of our network is shown in Figure 8. 

\begin{figure*}[hbt]
	
	\label{fig:8 Total Training Procedure}
	\centering
	\includegraphics[scale=.3]{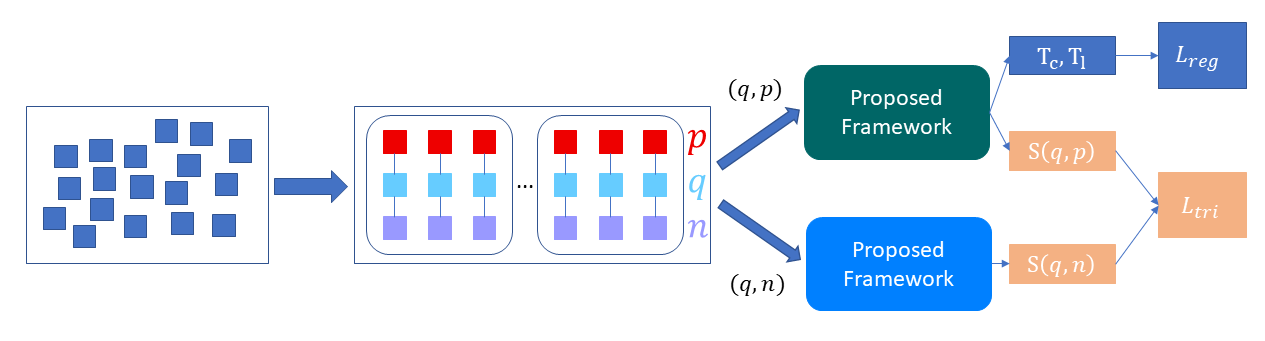}
	\caption{
		The total training procedure of our proposed network. the triplet $(q,p,n)$ are first sampled from training dataset, where $(q,p,n)$ denotes query, positive and negative samples in one triplet.  Then the triplet is provided to our model to obtain a score which indicates the similarity between the two video pairs. The scores of positive and negative video pairs are both used to train the network with the triplet loss function. Only query and positive video pairs are used to compute the regression loss.
	}
\end{figure*}

\subsubsection{Data Preparation}

For training set, we use triplet $(q, p, n)$ as input. The reason of using triplet can be shown in Figure 9, we want to get features from same class closer ("with high similarity score"), and push features from different classes further ("with low similarity score"). To generate one triplet, we first pick out one video in training set as query video. Then we pick out one ground truth section as query clip. Then we pick out 2 videos: one is with the same semantic label as query video clip, which is regarded as positive video; and another is with different semantic label as query video clip, which is regarded as negative video. For positive and negative video, we use Temporal Actionness Grouping (TAG) \cite{49_tag_proposal} to get many proposals. Then, for each video, we picked out the proposal with the highest intersection of union with its corresponding ground truth sections as our positive/negative proposal. Then we have the query video clip, positive proposal, negative proposal, which is $(q, p, n)$ denoted above. 

\begin{figure*}[hbt]
	\label{fig:9}
	\centering
	\includegraphics[scale=.3]{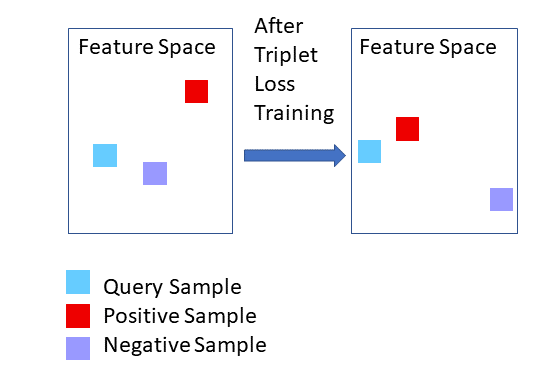}
	\caption{
		The reason for training triplet loss in our method. We want the videos’ visual feature similarity to be the same as the similarity of semantic label. Triplet loss‘s training procedure pushes the dissimilarity visual features away, and pulls the similarity visual features near, which meets our need.  
	}
\end{figure*}

For test set, we use pair $(q, p)$ as input, which implies that we do not use negative sample while testing. The query video clip's generation method is the same as that in training phase. Then we pick out the video with the same semantic label as positive video. Then we also use TAG to get many proposals, which is the same as generation positive proposal in training phase. Different from picking one out in training phase, we store all of the proposals in test phase, which can be written as $p=(p_1, ..., p_m)$. The reason is that we want to mimic the pratical use, since we do not know the ground truth in practice. Then we have query video and positive proposals, which is $(q, p)$ denoted above.

\subsubsection{Network Settings} 

For both datasets, we use pretrained C3D feature as input, which has a temporal resolution of 16 frames. And we use PCA to reduce output C3D feature dimension to 500, which follows the dataset settings in \cite{18_video_reloc}. In dataset, the video lengths of query video clips and proposal are likely to be different, which makes it inconvenient represent a batch of weighted graph adjacency matrices. In order to solve this problem, we set both query video clips and proposal’s total timesteps $T$ to 40: if a video’s timestep is more than 40, we select 40 equidistance features instead. And if a video’s length is less than 40, we use zero-padding to pad the remaining timestep, so that the total timestep of this video is also 40, just like what we have mentioned in section. 

We use PyTorch 1.4.0 to implement our proposed method. Our batch size is set to be 32. We train our proposed method for 10000 epochs. For triplet loss $L_{tri}$, an Adam optimizer \cite{51_adam} with $learning\_rate=1e-4$, $\beta_1=0.9$ and $\beta_2=0.999$ is applied to optimize this loss. For regression loss $L_{reg}$, we only train this loss’s gradient with respect to parameters only in the regression module. Another Adam optimizer with $learning\_rate=0.1$ is set to optimize regression loss. What’s more, we use dropout \cite{dropout} in LSTM module \cite{lstm} mentioned in section to increase its robustness. 

As for our score and regression module, the detailed implementation is shown in Figure 10. Note that the most significant difference between two modules is the output layer: score module only has one output, which represent the similarity score between two videos, while regression module has two outputs, which represents the starting and ending point of the proposal. In these two modules, batch normalization is applied after every convolution layer except the final one so that the features can be regularized. Another small difference in the two module is that in score module, the output $s$ in Score Module is restricted to $[-1, 1]$, while the output $[T_c, T_l]$ does not have this restriction, and $[T_c, T_l]\in\mathbb{R}^2$. 

\begin{figure*}[hbt]
	
	\label{fig:10}
	\centering
	\includegraphics[scale=.3]{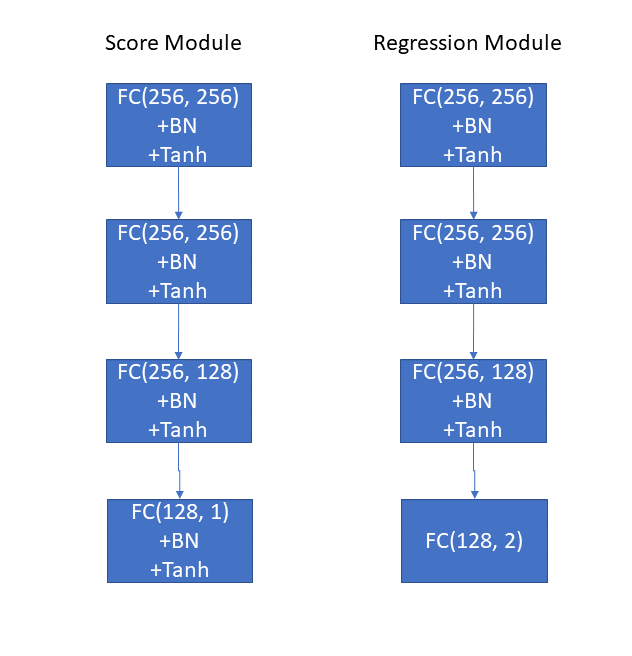}
	\caption{
		The detail of Score Module and Regression Module. 
		FC denotes Fully Connected Layer, BN denotes Batch Normalization Layer, Tanh denotes tanh function. 		 
	}
\end{figure*}

\subsubsection{Evaluation Metrics}

For video relocalization task, the evaluation metric is mean Average Precision (mAP) with a threshold temporal Intersection of Units (tIoU). The predicted segment in the reference video is considered correct if the tIoU with its ground truth instance is greater than the tIoU threshold. In our experiments, we use mAP@1 as our evaluation metric. (In the following paragraph, we will use mAP to denote mAP@1) 

The calculation of mAP can be written as follows:

$$ \mathrm{mAP} = \frac{1}{N} \sum_{i\in D_{test}} \sum_{j\in D_{test, i}} 1_{\mathrm{tIoU}_j > \mathrm{tIoU}_{th}} $$

where $N$ denotes the total number of query video clips in test set. $D_{test}$ denotes test set, $D_{test, i}$ means the i-th class in test set. $1_{\mathrm{tIoU}_j > \mathrm{tIoU}_{th}}$ means the value is 1 if and only if the the proposal of j-th query is higher than the threshold tIoU, otherwise, the value is 0. 
For our experiment, the evaluated tIoU thresholds  are {0.5, 0.6, 0.7, 0.75, 0.8, 0.85, 0.9, 0.95} in ActivityNet, and {0.5, 0.6, 0.7, 0.75, 0.8, 0.85, 0.9, 0.95} in Thumos14.
The threshold step in ActivityNet is 0.1 at first and then change to 0.05. The reason is that we found that the result changes slightly when tIoU threshold is lower than 0.7.

\subsection{Baselines and Results}

\subsubsection{Baselines}

We use the methods in \cite{18_video_reloc} and \cite{ruolin} as baseline methods. 

In \cite{18_video_reloc}, we use Frame Level Baseline, Video Level Baseline, SST and Cross Gated Bilinear Matching Baseline as our baseline methods. And in \cite{ruolin}, we use AFT+SRM mentioned in this article as baseline. 

Frame Level Baseline. This baseline ins proposed in \cite{18_video_reloc}, and it is similar to the approach in \cite{54_Chou2015Pattern}. After normalizing the query video's and reference video's features, a distance table $D_{ij} = \Vert h_i^q - h_j^r \Vert _2^2$ is calculated to represent the similarity between each timestep in query video and reference video. The diagonal block with the smallest average distances searched by dynamic programming. The output of this method is the segment in which the diagonal block lies. Note that no training is needed for this baseline. 

Video Level Baseline. This baseline is proposed in \cite{18_video_reloc}. In this baseline, they use LSTM as the video encoder and the L2-normalized last hidden state in LSTM is selected as video representations for video segments. Different from our method, their choice of positive and negative video clip is directly related to temporal Intersection over Unions (tIoU) in one reference video instead of two. The positive video clip has tIoU over 0.8, while the negative video clip's tIoU is less than 0.2. 

SST Baseline. \cite{53_sst} proposed Single-Stream Temporal Action Proposals (SST) method. In their training stage, they also use triplet as training input. We train the SST model on training set, and test on test set. The output of SST module is the proposal with the highest confidence score. 

Cross Gated Bilinear Matching BaseLine. \cite{18_video_reloc} proposed this method for video relocalization task. In their method, every timestep in the the problem is formulated as a sequence labelling problem: giving a query video and a reference video. The aim is to predict the classes of each timestep in reference video as one of $[start, end, in, out]$, where "start" denotes the starting point of a proposal video clip, "end" denotes the ending point of a proposal video clip, "in" denotes inside a proposal video clip, "out" denotes outside a proposal video clip. 

AFT+SRM Baseline. \cite{ruolin} uses Attention-Based Tensor and Semantic Relevance Module to make. They use triplet $(q, p, n)$ as input. And the choice of positive and negative videos is the same as our method. 

\subsubsection{Results on ActivityNet dataset}

The comparison result is listed in Table 1. From the table, we can see that our proposed method has an advantage over the existing video relocalization methods. For low tIoU thresholds (tIoU=0.5), it increases a lot (compared with Cross Gated Bilinear Matching, our method raises 3.48\% when tIoU=0.5; compared with AFT+SRM, our method raises 8.68\% when tIoU=0.5). In high tIoU thresholds (tIoU=0.9, tIoU=0.95), compared with Cross Gated Bilinear Matching, ours mAP raises 14.77\% when tIoU=0.9; compared with AFT+SRM, ours mAP raises 5.97\% and 2.42\% when tIoU=0.9 and 0.95). In all, the evaluation metric raises in all tIoU thresholds. 

\begin{table}
	\centering
	\caption{Results on ActivityNet dataset}
	\begin{tabular}{|l|l|l|l|l|l|l|l|l|}
		\hline
		Methods$\backslash$tIoU & 0.5 & 0.6 & 0.7 & 0.75 & 0.8 & 0.85 & 0.9 & 0.95 \\ \hline
		Chance & 16.10\% & 11.30\% & 5.60\% & - & 3.10\% & - & 1.20\% & - \\ \hline
		Frame-Level Baseline & 20.20\% & 14.60\% & 10.40\% & - & 5.40\% & - & 2.50\% & - \\ \hline
		Video-Level Baseline & 25.40\% & 18.10\% & 12.70\% & - & 6.30\% & - & 2.60\% & - \\ \hline
		SST & 34.70\% & 25.80\% & 18.30\% & - & 8.10\% & - & 3.00\% & - \\ \hline
		Cross-Gated Bilinear Matching & 45.80\% & 37.70\% & 28.20\% & - & 17.10\% & - & 7.30\% & - \\ \hline
		AFT+SRM & 40.60\% & 40.50\% & 40.40\% & 35.50\% & 30.00\% & 24.10\% & 16.10\% & 7.50\% \\ \hline
		Ours & 49.28\% & 49.05\% & 49.30\% & 44.01\% & 37.46\% & 29.59\% & 22.07\% & 9.92\% \\ \hline
		Ours(CNN) & 47.82\% & 47.64\% & 46.66\% & 42.94\% & 36.03\% & 28.27\% & 20.21\% & 8.30\% \\ \hline
	\end{tabular}
\end{table}

\subsubsection{Results on Thumos14 dataset}

To further prove the effectiveness of our purposed method, we also conduct experiments on Thumos14 dataset. The implementation details are the same as those in ActivityNet. And the result is listed in Table 2. From the result, we can find that our method still outperforms other methods in Thumos14 dataset. What’s more, we can find that the pattern we found in ActivityNet datastet still applies in Thumos14 dataset (the descending trend appears as tIoU gets larger). With the same threshold tIoU, the result in Thumos14 is lower than that in ActivityNet. We thought this is because these two datasets have some differences. 

\begin{table}
	\centering
	\caption{Results on Thumos14 dataset}
	\begin{tabular}{|l|l|l|l|l|l|}
		\hline
		Methods$\backslash$tIoU & 0.5 & 0.6 & 0.7 & 0.8 & 0.9 \\ \hline
		Chance &  &  &  &  &  \\ \hline
		Frame-Level Baseline &  &  &  &  &  \\ \hline
		Video-Level Baseline &  &  &  &  &  \\ \hline
		SST &  &  &  &  &  \\ \hline
		Cross-Gated Bilinear Matching &  &  &  &  &  \\ \hline
		AFT+SRM & 36.50\% & 36.40\% & 36.20\% & 24.70\% & 9.80\% \\ \hline
		Ours & 42.11\% & 41.76\% & 40.02\% & 28.34\% & 11.01\% \\ \hline
		Ours(CNN) & 40.49\% & 40.12\% & 38.75\% & 27.31\% & 9.29\% \\ \hline
	\end{tabular}
\end{table}

\subsubsection{Qualitative results}

Also, we show the qualitative results of our proposed method to demonstrate the effectiveness of our method intuitively. We picked out 2 classes from ActivityNet dataset (i.e., Cleaning Windows and Tai-Chi) and two classes from Thumos14 dataset. The results are shown in Figure 11. It can be seen that ground truth and our proposed method overlap a lot. Although our proposed module has not seen test classes before, it can effectively measure the semantic similarities between query and reference classes. 

\begin{figure*}[hbt]
	
	\label{fig:10}
	\centering
	\includegraphics[scale=.3]{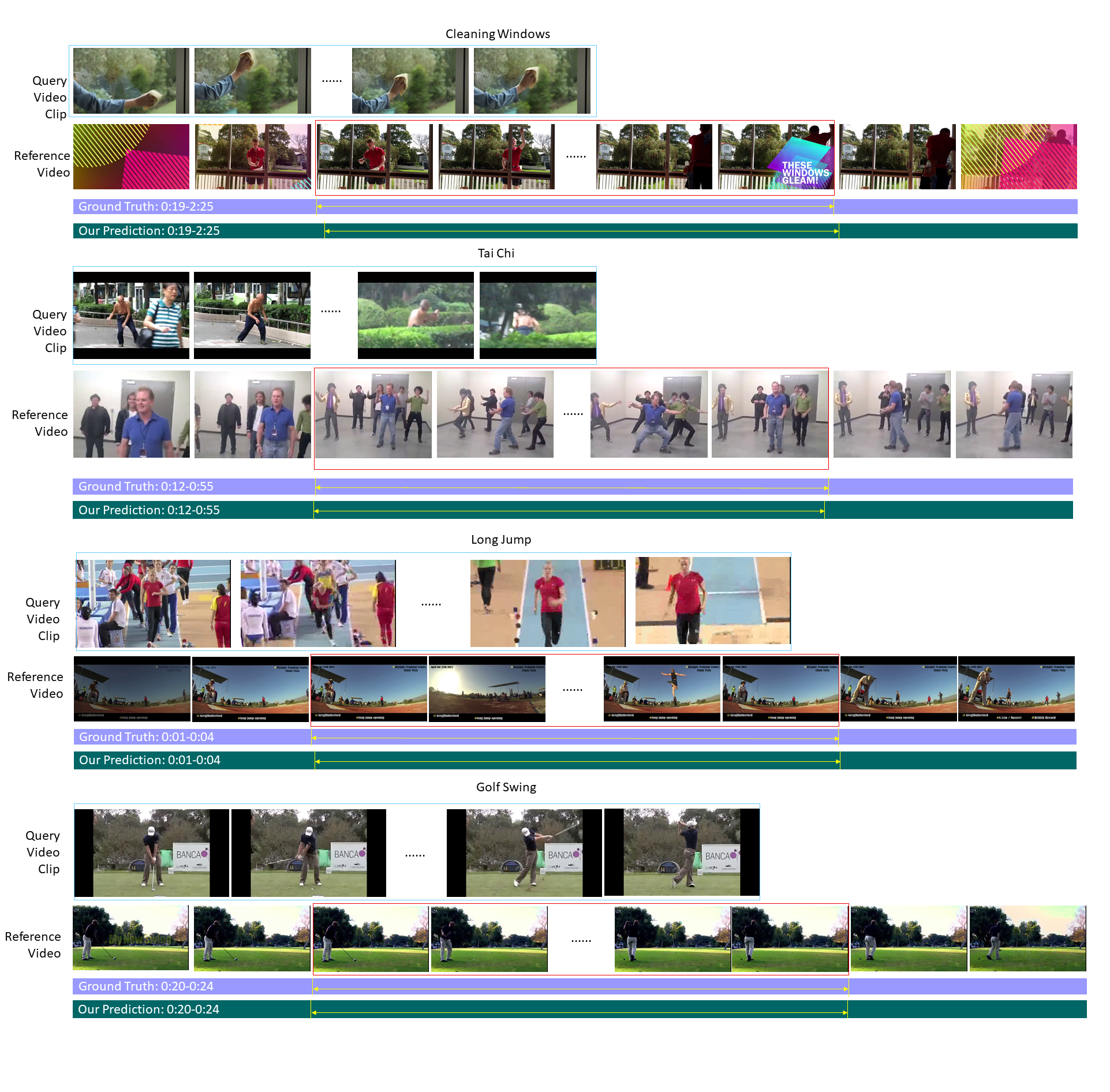}
	\caption{
		Quality result of our video moment retrieval method on ActivityNet dataset and Thumos14 dataset. We pick out “Cleaning Windows” and “Taichi” from AcitivityNet, and “Long Jump” and “Golf Swing” from Thumos14 dataset. 
		Query is the query video, reference is the video which has the same semantic label with query video. 
	}
\end{figure*}

\subsection{Ablation Study}

Next, we conduct ablation study to find out the best combination of Multi-Graph Feature Fusion block on number of Multi-Graph Feature Fusion layers (n) and different kinds of graphs in one Multi-Feature Fusion layer (k). And we want to find out if graph information works in our method. We perform ablation study on ActivityNet v1.2 dataset, and we list all the results in Table 3. 

\begin{table}
	\centering
	\begin{tabular}{|l|l|l|l|l|l|l|l|l|}
		\hline
		& 0.5 & 0.6 & 0.7 & 0.75 & 0.8 & 0.85 & 0.9 & 0.95 \\ \hline
		n=1, k=1 & 48.62\% & 47.89\% & 48.63\% & 42.19\% & 35.61\% & 28.21\% & 21.05\% & 9.38\% \\ \hline
		n=1, k=2 & 48.74\% & 48.21\% & 48.99\% & 43.22\% & 36.64\% & 28.53\% & 21.47\% & 9.44\% \\ \hline
		n=1, k=3 & 46.87\% & 46.29\% & 47.10\% & 42.23\% & 34.57\% & 27.85\% & 21.05\% & 9.30\% \\ \hline
		n=2, k=1 & 49.18\% & 48.42\% & 48.74\% & 42.90\% & 37.27\% & 30.27\% & 21.41\% & 8.71\% \\ \hline
		n=2, k=2 & 49.22\% & 48.42\% & 48.61\% & 43.87\% & 37.39\% & 28.92\% & 21.91\% & 8.84\% \\ \hline
		n=2, k=3 & 49.28\% & 49.05\% & 49.30\% & 44.01\% & 37.46\% & 29.59\% & 22.07\% & 9.92\% \\ \hline
		n=3, k=1 & 43.10\% & 42.96\% & 42.69\% & 37.03\% & 30.61\% & 24.90\% & 15.81\% & 6.80\% \\ \hline
		n=3, k=2 & 45.99\% & 44.48\% & 44.61\% & 40.56\% & 30.76\% & 25.20\% & 17.38\% & 7.67\% \\ \hline
		n=3, k=3 & 44.02\% & 43.63\% & 44.39\% & 39.94\% & 32.76\% & 23.70\% & 18.05\% & 6.83\% \\ \hline
	\end{tabular}
\end{table}

\subsubsection{Number of layers}

We evaluated different depths (n) of our graph convolution module to find out which depth is best. In the experiment, we tested on n=1, 2, 3. Table 3 illustrates the result of this part. We can find that n=2 is the best among all these depths. n=1 is slightly worse than n=2 (in tIoU=0.9 and k=3, n=1’s result is 21.05\%, while n=2’s is 22.07\%). But n=3’s result is much lower than that of n=2 and n=1 (in tIoU=0.9 and k=3, n=3’s result is 18.05\%, which is much lower than 21.05\% and 22.07\%). The reason is that Graph Convolution Network is a kind of Laplacian smoothing procedure, i.e., as module goes deeper, the nodes’ features are smoothed to be the same, so the features of different nodes degenerate into the features of the whole graph. In our task, we need these nodes to represent different timesteps’ features, so features in different timesteps should not be smoothed to be the same.

\subsubsection{Number of scales in one layer}

Also, we evaluated the results in using different kinds of graphs (k) in one Multi-Graph Feature Fusion layer. In this experiment, we tested on k=1, 2, 3. Table 3 illustrates the result of this part. We can see that k=3 is the best among all the ks (for example, when tIoU=0.95, k=3’s result is 1.21\% and 1.08\% higher than k=1 and k=2 respectively), which implies that using more kinds of graph connections in one layer can extract more kinds of feature, and fusing them can have a better result. What’s more, observing the increase trend from k=1 to k=2 and from k=2 to k=3, we find out that the increase trend is becoming slower as k goes bigger, so we treat k=3 as the most balancing result.

\subsubsection{The effectiveness of graph convolution}

To show the necessity of using graph convolutions, we also conduct an experiment on Multi-Graph Feature Fusion Layer. As the ablation study on n and k shows, we use $n=2, k=3$ as our benchmark in the following ablation study. As the graph convolution operation shows in section 3, a graph convolution consists of 2 steps: feature aggregation via weighted adjacency matrix and feature projection via trainable weights. To prove if the effectiveness of Multi-Graph Feature Fusion Layer is to prove the effectiveness of feature aggregation in step 1, which implies that there only leaves feature projection, which is the same as original fully-connected layer (or convolutional layer). 

In practice, we use 3 $1\times1$ convolutions instead of 3 graph convolutions as comparison in one layer. The result is also shown in Table 1 and Table 2 (Ours v.s. Ours(CNN)). From the comparison, we can see that with multi-graph information, we can have a better result in test stage. 

\section{Conclusion}

In this article, we found out a phenomenon which shows that there is no consistent relationship between feature similarity by the frame with same timestep in the two videos and feature similarity by the two videos overall. Then, to alleviate this phenomenon, we proposed a new module called Multi-Graph Feature Fusion Module: we first treat query and proposal video clips’ features as graphs, where each timestep is treated as nodes, and we build weighted graph adjacency matrices of this graph. Then we use Multi-Graph Feature Fusion Module to extract the graph’s feature. Extensive experiments results show that our proposed method is effective, and outperforms the baseline methods in ActivityNet v1.2 dataset and Thumos14 dataset.

\bibliographystyle{ieeetr} % ieee trans style
\bibliography{my_reference.bib} % my references

\end{document}